\documentclass{article}
\usepackage{amssymb}

\usepackage[preprint]{corl_2025} 

\usepackage{todonotes}
\usepackage{caption}
\usepackage{wrapfig}
\usepackage{float}
\usepackage{amsmath}
\usepackage{amssymb}
\usepackage{mathtools}
\usepackage{amsthm}
\usepackage{caption}
\usepackage{subcaption}

\usepackage{algorithm}
\usepackage{algorithmic}
\usepackage{amsfonts}       

\title{Zero-Shot Visual Generalization in Robot Manipulation}

\author{
  Sumeet Batra\\
  University of Southern California \\
  United States\\
  \texttt{ssbatra@usc.edu} \\
  \And
  Gaurav S. Sukhatme \\
  University of Southern California \\
  United States \\
  \texttt{gaurav@usc.edu} \\
}

\begin{document}
\maketitle


\begin{abstract}
    Training vision-based manipulation policies that are robust across diverse visual environments remains an important and unresolved challenge in robot learning. 
    Current approaches often sidestep the problem by relying on invariant representations such as point clouds and depth, or by brute-forcing generalization through visual domain randomization and/or large, visually diverse datasets.
    Disentangled representation learning -- especially when combined with principles of associative memory -- has recently shown promise in enabling vision-based reinforcement learning policies to be robust to visual distribution shifts. 
    However, these techniques have largely been constrained to simpler benchmarks and toy environments. 
    In this work, we scale disentangled representation learning and associative memory to more visually and dynamically complex manipulation tasks and demonstrate zero-shot adaptability to visual perturbations in both simulation and on real hardware. 
    We further extend this approach to imitation learning, specifically Diffusion Policy, and empirically show significant gains in visual generalization compared to state-of-the-art imitation learning methods.
    Finally, we introduce a novel technique adapted from the model equivariance literature that transforms any trained neural network policy into one invariant to 2D planar rotations, making our policy not only visually robust but also resilient to certain camera perturbations. 
    We believe that this work marks a significant step towards manipulation policies that are not only adaptable out of the box, but also robust to the complexities and dynamical nature of real-world deployment. Supplementary videos are available at \url{https://sites.google.com/view/vis-gen-robotics/home}.
\end{abstract}



\section{Introduction}

A key requirement of any generalist robot system deployed in the real-world is the ability to perform tasks across visually diverse environments. 
High-dimensional inputs like RGB images offer rich information but also introduce complexity due to the curse of dimensionality.
Given the enormous diversity of real-world visual data, accounting for every possible variation within a fixed dataset is intractable. 
Extracting the underlying structural knowledge of the world from visual data while being robust to semantically irrelevant visual perturbations remains an open question. 
The robot learning field has largely relied on one of several trends, one of which is to train agents in simulation, where visual complexity can be controlled and large-scale synthetic and diverse data can be generated efficiently through GPU-accelerated simulators~\citep{maniskill, ige, gen2sim}.
However, transferring policies trained in simulation to the real world is hindered by the "Sim2Real" gap caused by mismatches in fidelity and unmodeled dynamics.
Domain randomization is the leading strategy to close this gap by varying the simulation parameters such that real-world conditions fall within the distribution of the training data.
Domain randomization has proven effective in both simulated benchmarks and real-world robotic tasks when the data diversity is sufficiently large~\cite{openai_dextrous, svea, openai_rubics}.
A seemingly unrelated but conceptually similar approach to visual generalization in the age of foundation models has been to train large models, typically Vision-Language-Action (VLA) models, on very large real world robot datasets \cite{openvla, rt1, rt2, fast}.
By learning from varied contexts, these models generalize to novel environments, mirroring the principles of domain randomization at scale.

\begin{figure}[t]
    \centering
    \includegraphics[width=1.0\linewidth]{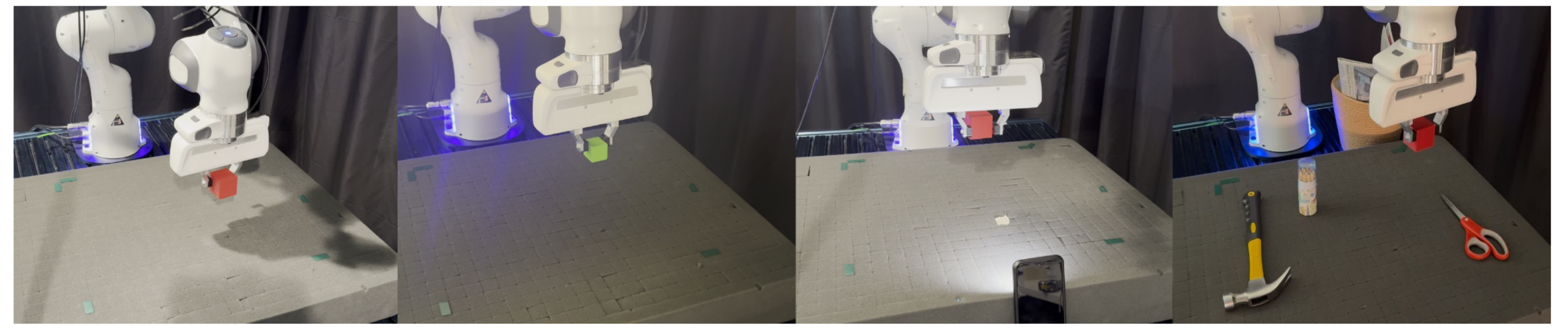}
    \caption{Behavior cloning with disentangled representations and associative latent dynamics achieves zero-shot generalization to various real world perturbations, such as changes in ambient lighting (\textbf{left}), object color (\textbf{middle-left}), directed lighting (\textbf{middle-right}), and the presence of distractor objects (\textbf{right}).}
    \label{fig:main_fig}
\end{figure}

Despite recent advances, whether these models are truly capable of \textit{extrapolative generalization} remains questionable.
For example, \cite{neural_foundations} showed that vision-language models trained for pixel-wise future prediction perform poorly on out-of-distribution (OOD) data.
Similarly, \citep{nn_unusual_pose} found that state-of-the-art classifiers suffered up to a 30\% accuracy drop when tested on objects in unusual poses. 
While one might mitigate this with orientation randomization, doing so would require exhaustively covering every possible orientation of every object we wish to classify -- a clearly impractical solution.
Compounding the issue, \cite{nn_symmetries_data} shows that neural networks trained via supervised learning often fail to extract and generalize fundamental symmetries such as SO(2) rotations. 
This suggests that, despite all the efforts on exhaustive visual domain randomization in simulation or collecting massive and diverse real-world datasets, current models may not truly generalize and that data scaling, while important, is insufficient. 
Despite these shortcomings, it is well known that biological systems robustly extract structure from high-dimensional sensory inputs, even under severe visual perturbations~\cite{brain_solve_recognition, behrens2018cognitive}. The neuroscience literature points to \textit{factorized, modular} representations as a key enabler of this kind of structural generalization~\citep{behrens2018cognitive, bakermans2025constructing, tang2025hippocampal}. In machine learning, similar principles appear in disentangled and object-centric representation learning~\cite{bVAE, qlae, slot_ae}, which have been shown to facilitate visual generalization in continuous control tasks in simulation \cite{sac_ae, darla, alda}. However, these approaches have yet to scale beyond toy problems or narrow benchmarks.

A recent approach that shows promise in scaling to harder tasks is Associative Latent DisentAnglement (ALDA)~\citep{alda}, a reinforcement learning (RL) algorithm that learns factorized representations via disentanglement and leverages principles of \textit{Associative Memory} to achieve SOTA performance on a popular continuous control and visual generalization benchmark~\citep{hansen2021stabilizing}.
Given an OOD observation at test time, ALDA decomposes the observation into a disentangled latent representation and maps back \textit{specific} dimensions that are OOD to in-distribution values, equivalent to recalling the most related observation it \textit{has} seen and taking an action based on that instead. 
This work investigates whether ALDA's principles can extend to robotic manipulation and real-world deployment, where visual generalization remains challenging.
While ALDA shows promise, RL alone struggles with complex manipulation tasks and cannot be directly trained from real-world interaction due to poor sample efficiency, making imitation learning a more practical alternative.
We therefore extend ALDA to imitation learning, specifically diffusion-based behavior cloning  methods~\citep{diffusion_policy, 3d_diff_policy, 3d_diffuser}, and find similar gains in zero-shot visual generalization.
In addition to visual variations, real-world deployments can face camera perturbations, which further degrade performance.
To address this, we draw on Equivariant Neural Networks, which encode symmetry structures (e.g., rotations) into their architecture and have shown strong generalization and sample efficiency in robot learning~\citep{so2_equiv_rl, equidiff} on robot learning tasks. 
However, these methods require training from scratch, significantly increasing training time and imposing constraints on the available model architectures. 
As an alternative, we introduce \textit{learned canonicalization}, adapting recent work \cite{equiadapt, equi_learn_canon} to the robot learning context. 
This family of methods uses a lightweight, surrogate equivariant neural network to \textit{finetune} larger pretrained models to become equivariant to certain symmetries (e.g., SO(2) transformations) -- without retraining from scratch. 
However, these methods have only been studied within simpler supervised learning contexts such as image classification and segmentation.
We adapt this approach to robot learning and present an algorithm for turning any pre-trained robot policy into an equivariant one immune to discrete planar rotations in the SO(2) group. 

To summarize, our contributions are as follows: \textbf{(1)} We evaluate ALDA on visually rich and dynamically complex manipulation tasks and demonstrate strong visual generalization, \textbf{(2)} we extend ALDA to imitation learning, specifically diffusion-based behavior cloning, and demonstrate similar gains in generalization performance over SOTA imitation learning baselines, \textbf{(3)} we propose a finetuning method to make any pretrained robot policy equivariant to discrete planar rotations, and \textbf{(4)} we validate our model on a real robot under realistic visual perturbations. 
By integrating these methods, we take a significant step towards generalist real-world agents capable of robust, zero-shot adaptation across lighting changes, background clutter, and camera perturbations.

\section{Related Work}
\textbf{Reinforcement Learning for Robotics.}
RL algorithms learn a policy that maximizes the discounted sum of future rewards according to a given reward function.
On-policy methods \cite{ppo, trpo, a3c} learn from experience collected by the behavior policy, while off-policy methods \cite{sac, ddpg, td3, dqn} can learn from prior experience. 
Model-based RL algorithms construct explicit predictive models of the environment, which are used for planning, and can learn from imagined experience \cite{dreamer, planet, tdmpc2, ha2018world}. 
These methods have been shown to be effective at solving tabletop manipulation tasks~\citep{tdmpc2}, contact-rich assembly tasks~\citep{industreal, automate}, and dexterous manipulation~\citep{openai_dextrous, dextreme} on modern simulators~\citep{maniskill, ige, dmcontrol} and on real robots.

\textbf{Learning from Demonstrations.} Due to the sample inefficiency of RL and the complexity of certain tasks, imitation learning has become an increasingly popular paradigm that allows robots to leverage sparser real-world, expert demonstration datasets~\citep{what_matters_in_il, teach_robot_fish}.
Generative methods such as diffusion models \cite{ddpm, ddim, diffusion_song, high_res_diff} have shown success at solving manipulation tasks where the data distribution exhibits multiple modalities~\cite{diffusion_policy, 3d_diff_policy, 3d_diffuser, diffusion_dagger, diff_imitate_human}.
Recently, Transformer~\citep{vaswani2017attention}-based methods have shown dramatic improvements in generalization across tasks~\cite{peract, rvt, act}.
In a similar vein, there has been a recent trend to train "Robot Foundation Models", using either from-scratch Transformer models or pre-trained Vision-Language Models (VLMs) converted into Vision-Language-Action (VLA) models by training on large demonstration datasets~\citep{rt1, rt2, openvla, fast}.  
However, there is a growing body of evidence~\citep{nn_symmetries_data, nn_unusual_pose, neural_foundations} to suggest that LLMs, VLMs, and more broadly current Deep Learning architectures, are not achieving systematic extrapolative generalization, implying that the downstream models finetuned for robotics tasks likely have gaps in performance and many edge cases despite claims of generalization. 

\textbf{Representation Learning for Control.} 
We concur with and adopt the position taken by several recent works~\citep{neural_foundations, lecun_intuitive_phys, vinn} which argue that structured representations are crucial to extrapolative generalization in agentic tasks like future prediction and planning. 
From the cognitive and neuroscience literature, there is mounting evidence that biological agents are capable of rapid adaptation and generalization in part thanks to modular, structured representations \cite{behrens2018cognitive, brain_solve_recognition, tang2025hippocampal, samborska2022complementary, sun2023organizing}, which we posit will greatly aid artificial agents in doing the same. 
One such approach to learning structured representations in machine learning is through disentangled representation learning (DRL)~\citep{bVAE, bio_ae, qlae, fsq}, in which a model learns a latent representation of high-dimensional data (e.g., images), where each dimension represents one factor of variation (e.g., object color, size, shape, etc.). 
DRL is thought to be a key component of compositional generalization~\citep{disentangle_comp_gen, behrens2018cognitive}, and has shown promising results in visual generalization in continuous control tasks from high-dimensional image observations~\citep{alda, darla, sac_ae}.
ALDA~\citep{alda}, which we use as the foundation for this work, combines learning factorized representations with principles of associative memory from modern, continuous Hopfield Networks~\citep{hopfield1982neural, hopfield_all_you_need, energy_hopfield}.
Hopfield networks store memories as fixed points and use attractor dynamics to recover the most similar memory given an input query. 
Recent work in neuroscience finds evidence that the hippocampus in mice achieves rapid generalization through disentangled memory representations~\citep{tang2025hippocampal}, providing a biologically plausible motivation for this line of work.


\textbf{Equivariance via Learned Canonicalization.} 
Learned canonicalization is a method by which a larger pretrained network, called the predictor network $p$, can be made equivariant using a smaller, surrogate equivariant neural network. 
An equivariant function $h$ is one that commutes with a \textit{group action} $\rho(g)$ on a symmetry group $\mathcal{G}$ such that $h(\rho(g) \cdot \textbf{o}) = \rho'(g) \cdot h(\textbf{o})$ i.e. an equivariant transformation on the input \textbf{o} induced by $\mathcal{G}$ results in a predictable transformation of the  output $h(\textbf{o})$.
Normally, the predictor network would need to learn the inverse mapping from $h$.
However, one can instead learn a \textit{canonicalization network} $C: \mathcal{O} \rightarrow G$ that can transform elements of the input from their orbit into a \textit{canonical} sample where $h$ can be applied, and then transforming the sample back to its original position in the orbit:

\begin{equation}
    \label{canon}
    h(\textbf{o}) = \rho'(C(\textbf{o})) \cdot p \left( \rho (C(\textbf{o})^{-1} ) \cdot \textbf{o} \right)
\end{equation}

This formulation alleviates the burden of modifying the pretrained network $p$, instead putting it on the canonicalization network, which can be a lightweight equivariant neural network.
This can be especially useful when training robot policies where issues such as stabilizing learning and sample/time complexity are more relevant. 
Indeed, \cite{so2_equiv_rl} requires using equivariant neural networks for the RL agent itself, along with additional assumptions on the MDP that ensure SO(2) equivariance is not violated during training. 
Learned canonicalization has shown success on image classification and segmentation tasks~\citep{equiadapt, equi_learn_canon}, but has not been studied in the robot learning context.
To the best of our knowledge, we are the first to demonstrate how this method can be used to make vision-based robot policies robust to camera perturbations.

\section{Method}
\begin{figure}
    \centering
    \includegraphics[width=1.0\linewidth]{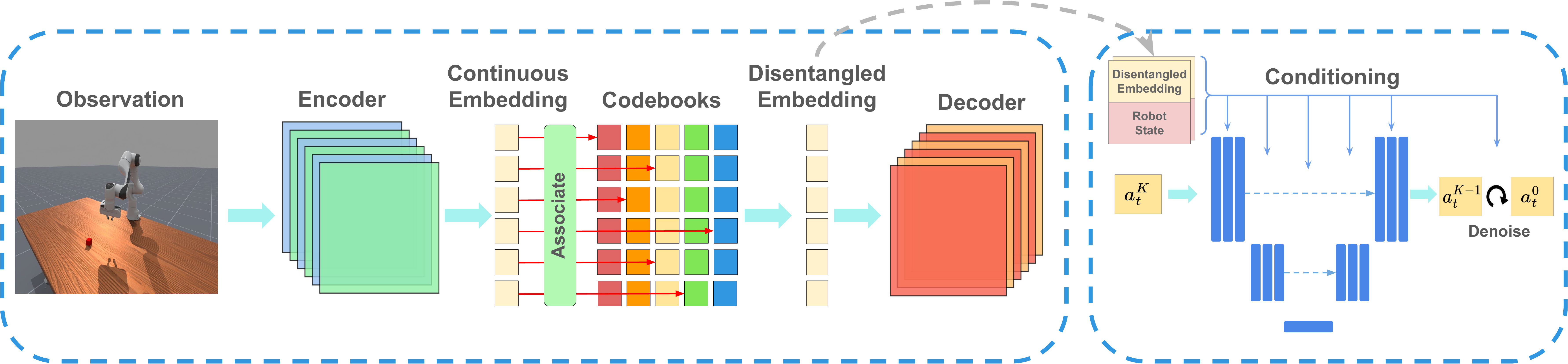}
    \caption{Overview of ALDA + Diffusion Policy (\textbf{ALDA-DP}). ALDA-DP jointly learns a factorized representation of the image observation while training the policy. The diffusion model denoises actions conditioned on this representation.}
    \label{fig:alda_dp_method}
\end{figure}

Diffusion models have shown impressive results on current manipulation benchmarks~\citep{3d_diff_policy, 3d_diffuser, diffusion_policy, diff_imitate_human, diffusion_dagger}; thus, we choose a diffusion-based actor as our driver.
We describe our approach to learning disentangled representations and leveraging principles of associative memory for Diffusion Policy~\citep{diffusion_policy}.
An overview of the method is presented in Figure~\ref{fig:alda_dp_method}.
From now on, we will refer to the Diffusion Policy variant as ALDA-DP, and the RL variant as ALDA-SAC. 
ALDA-SAC learns a factorized representation while jointly training an RL agent using Soft Actor-Critic~\citep{sac}, and assumes the same observation and action representations. 
We refer the reader to \citep{alda} and the Appendix for implementation details on ALDA-SAC. 

\subsection{Disentangling Representations for Behavior Cloning}
ALDA receives an image observation and encodes it into a continuous latent representation $z_{cont.} \in \mathbb{R}^{n_z}$ with the encoder $f_{\theta}$. 
$z_{cont.}$ is mapped to a discrete representation $z_d \in \mathbb{R}^{n_z}$ via a \textit{collection} of discrete scalar codebooks $Z = V_1 \times V_2 \times ... \times V_{n_z}$, one per latent dimension. 
The mapping is given by an associative latent dynamics model, which leverages the attention mechanism used in modern Hopfield networks to perform \textit{pointwise} attention between scalar values in $z_{cont.}$ and the codebooks to produce $z_d$:

\begin{equation}
    z_{d_j} = \text{Softmax} \left( \beta \text{Sim}(z_j, V_j) \right) \odot V_j
    \label{eq:associate}
\end{equation}

A decoder network $g_{\phi}$ uses $z_d$ to reconstruct the observation, thus propagating visual information back to $z_d$. 
The discrete nature of $z_d$  enforces separation and, combined with large network activation penalties, facilitates disentanglement within $z_d$.
More details about the associative latent mechanism and disentanglement can be found in the Appendix.
Of particular importance is that when presented with in-distribution observations at test time, the association step in \eqref{eq:associate} is approximately a no-op, because the encoder has been optimized to be close to the distribution $\textbf{z}_d$ and thus $\textbf{z}_{cont.}$ will likely already be $\textbf{z}_d$ or at least very similar. 
However, if the observations become out of distribution due to visual distractions, then Equation~\eqref{eq:associate} \textit{forces} the representation to be in-distribution before the actor model sees it. 
Unlike tasks with learnable modern Hopfield networks where the associations between two sets need to be learned explicitly, here, association is made possible without learning \textit{because} the representations are disentangled. 

\subsection{Training}
Diffusion-based policies learn to iteratively denoise actions conditioned on latent representations $z_t$ of image observations $\textbf{o} \in \mathcal{O}$ and robot proprioceptive state $\textbf{s} \in \mathcal{S}$.
Following prior works~\cite{diffusion_policy, 3d_diff_policy}, we assume access to and learn from a dataset of expert action trajectories $\{(\textbf{o}_1, \textbf{s}_1, \textbf{a}_1), (\textbf{o}_2, \textbf{s}_2, \textbf{a}_2), ..., (\textbf{o}_T, \textbf{s}_t, \textbf{a}_T) \}$, where $\textbf{o}_t$ is an image observation, $\textbf{s}_t$ robot proprioceptive state, and $\textbf{a}_t$ is a robot action.
Actions $a_t \in \mathbb{R}^4$ are delta \textit{xyz} position commands for the end-effector and a gripper open/close command $a_t = \{a_t^{\Delta_{\text{loc}}}, a_t^{\text{open}} \in \{ 0, 1\} \}$.
The actor predicts subsequences of future actions $\tau = a_{t:t+k}$ to a horizon of length $k$. 
We use a denoising probabilistic diffusion model~\citep{ddpm}, which is trained by iteratively adding noise according to a variance schedule $\beta_i$ to action subsequences and learning the inverse denoising procedure $p_{\theta}(\tau^{i-1} | \tau^i) = \mathcal{N}(\tau^{i-1}; \mu_{\theta}(\tau^i, i), \Sigma_{\theta}(\tau^i, i))$. 
During inference, the diffusion model, conditioned on the disentangled latent $\textbf{z}_d$ and robot proprioceptive state \textbf{s}, denoises a noisy action chunk $\tau^i$ into a noise-free action sequence $\tau^0$ that is executed by the robot. 
During training, we randomly sample a trajectory timestep $t$ and diffusion timestep $i$ and add noise $\epsilon$ to the ground truth action sequence $\tau^0$. 
We use mean-squared error (MSE) to predict the noise at $i$:
\begin{equation}
    J(DP) = ||\epsilon_{\theta}(\textbf{z}_d, \textbf{s}, \tau^i, i) - \epsilon||_2^2.
    \label{eq:bc}
\end{equation}
The final training objective is therefore $J(ALDA) + J(DP)$ (see the Appendix for details on $J(ALDA)$).

\subsection{Equivariant Adaptation}
\label{subsec:equi_adapt_method}
We now describe our technique to make pre-trained policies invariant to discrete image rotations in SO(2), based on the method presented in \cite{equiadapt}.
Assume we are given a trained policy $\pi(a|z)$, encoder $f$, and latent model $l$. 
An lightweight equivariant neural network is initialized as the canonicalization function $C$. 
To enable our policy to take optimal actions under camera rotations, we must make $f$ and $l$ equivariant to group actions on $\textbf{o}$ \textit{and} make $\pi$ equivariant to the resulting group actions on $z$. 
Therefore, the prediction function $p$ in equation \eqref{canon} we wish to optimize is $\pi(\cdot|l(f(\textbf{o})))$. 
The canonicalization method presented in \cite{equiadapt} was studied under the context of supervised learning and assumes the existence of a ground truth dataset. 
In our case, the trained actor and encoder networks and latent model are our oracles, so we duplicate and freeze their weights, and refer to them as $\pi^*, f^*, \text{ and } l^*$. 
Rather than a reconstruction loss over images as in prior works, we wish to "reconstruct" the optimal actions of $\pi^*(\cdot | f^*(\cdot))$ given the canonicalized sample $C(\textbf{o})$ of the original observation $\textbf{o}$.
Since our RL variant of ALDA under the hood is Soft Actor-Critic, which maintains a replay buffer, we can save and reuse the buffer as a dataset to sample i.i.d. transitions. 
For the ALDA-DP variant, we sample transitions from the expert demonstration dataset. 

Since we disable gradient flow through the latent model, we add the commitment loss term from ALDA's training objective here to keep the continuous outputs of $f$ and discrete embeddings of $l$ close to each other during the optimization procedure. 
Finally, following in step with prior work, we also utilize a canonicalization prior (CP) regularizer to ensure the canonicalized inputs match the original observation inputs as closely as possible. 
This is done by minimizing the KL-Divergence between the transformed distribution induced by $C$ and the original data distribution i.e. $\mathcal{L}_{prior} = -\mathbb{E}_{\textbf{o} \sim D} \left[ D_{KL} (\mathbb{P}_D || P_{C(\textbf{o})}) \right]$.
The final objective can then be written as

\begin{equation} \begin{aligned} || \pi(a|l(f(\textbf{o})) - \pi^*(a|l^*(f^*(\textbf{o}))||_2^2 + \beta \cdot \mathcal{L}_{prior} + \mathcal{L}_{commit} \end{aligned} \end{equation}
where $\beta$ is a hyperparameter controlling the regularization strength. 
Since prior learned canonicalization methods have struggled with continuous image rotations, we also restrict our approach to the space of discrete rotations $C_n$. 
However, the number of discrete bins $n$ can be increased for robustness to more degrees of rotational camera perturbations, albeit at the expense of computational complexity. 
Pseudocode for the finetuning procedure is in the  Appendix.

\section{Experiments}
\label{sec:exps}

\begin{figure}[h]
    \centering
    \includegraphics[width=1.0\linewidth]{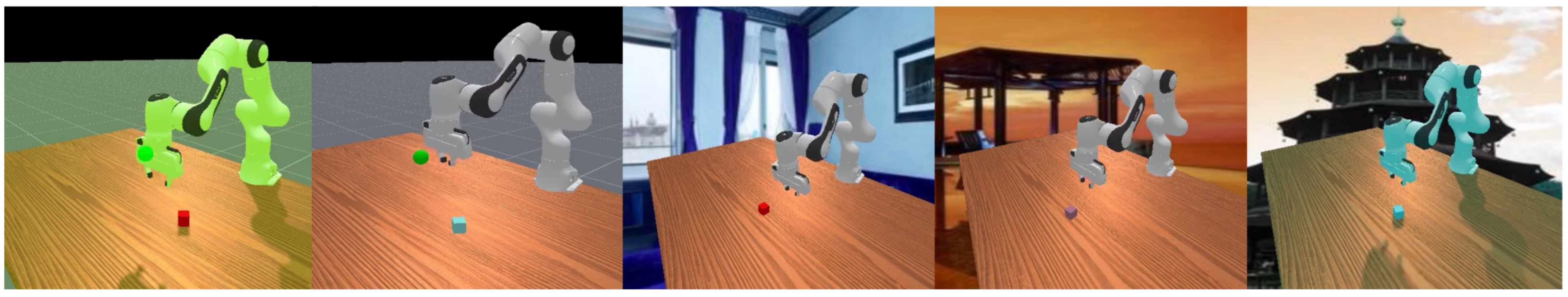}
    \caption{ManiSkill3 visual generalization tasks. \textbf{Left to right}: random lighting, random cube color, distracting background (DBG), DBG + random cube color, DBG + random lighting + random cube color.}
    \label{fig:mvgb}
\end{figure}

To investigate how our method scales to manipulation tasks, we choose ManiSkill3's~\citep{maniskill} tabletop manipulation suite. 
ManiSkill3 is a high-throughput simulator with support for GPU parallelization. It comes with demonstrations for imitation learning on many tabletop tasks out of the box, making it an obvious choice for benchmarking. 
Since we wish to test the robustness of our method and baselines to visual distribution shifts, we design a suite of visual scene randomizations on top of existing tasks, which we call the \textbf{ManiSkill Visual Generalization Benchmark} (MVGB). 
MVGB supports background scene randomizations, lighting intensity and lighting direction randomization, table color randomization, and object color/size randomization.
For this work, we benchmark on six sources of visual randomization, which are combinations of the following three principle randomizations: \textbf{(i) Distracting Backgrounds (DBG)} where we overlay an image randomly sampled from the Places365 dataset~\citep{zhou2017places} consisting of 1.8 million images from 365 different real-world environments, \textbf{(ii) Random Colors}, where the color of the object being manipulated is randomized, and \textbf{(iii) Random Lighting}, where the scene's ambient lighting color and intensity is randomized.

\begin{figure}[h!]
    \centering
    \includegraphics[width=1.0\linewidth]{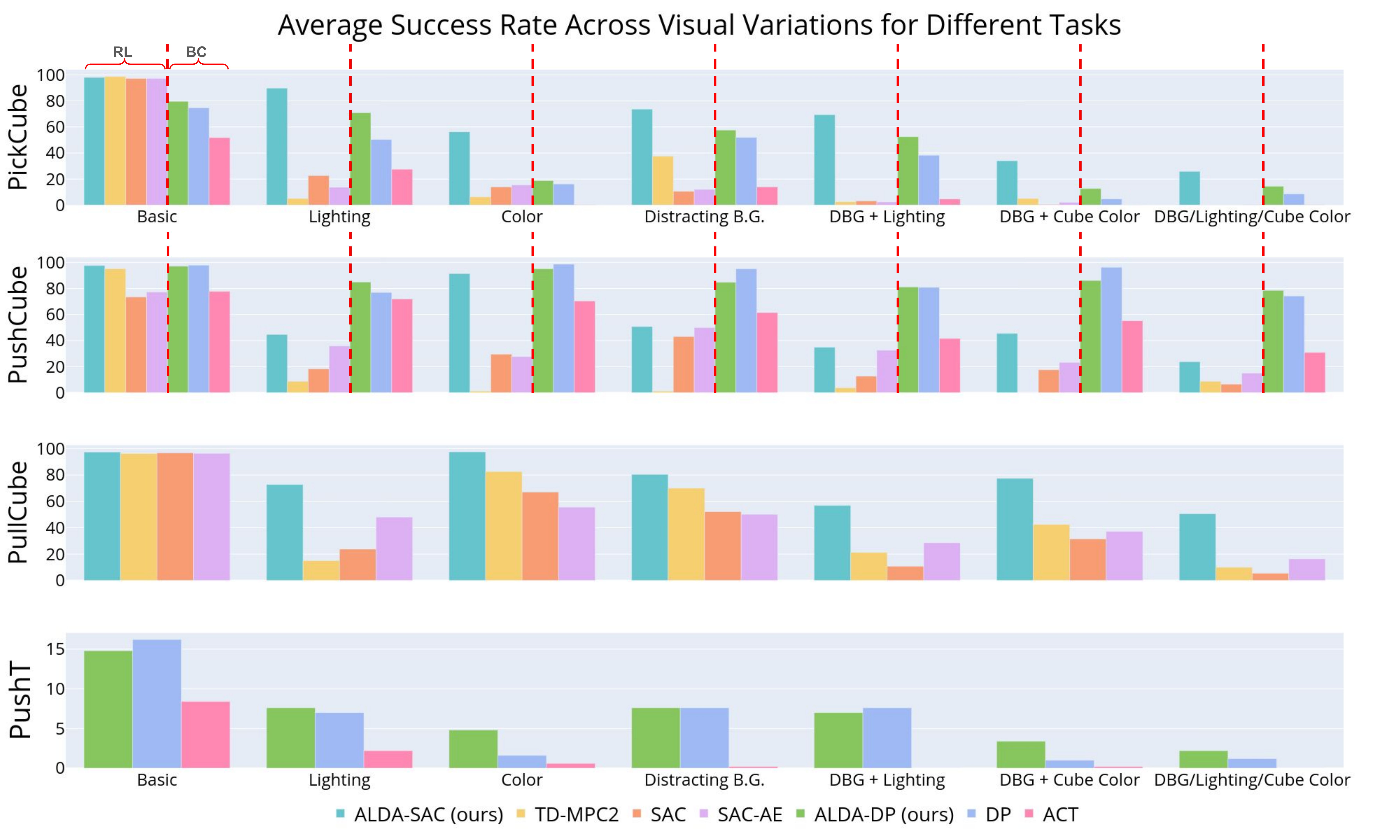}
    \caption{Average success rate of ALDA-SAC and ALDA-DP compared to various RL and BC baselines. The first two rows have both RL and BC results, while the 3rd row has only RL results and the fourth only BC results. ALDA-* methods overall perform the best, and with a large margin, especially on PickCube.}
    \label{fig:main_plots}
\end{figure}

We compare ALDA-SAC and ALDA-DP against various SOTA RL and BC baselines, respectively. 
For ALDA-SAC, we compare to \textbf{SAC}, \textbf{SAC-AE}~\citep{sac_ae}, which demonstrated promising generalization performance by training an autoencoder as an auxiliary objective, and \textbf{TD-MPC2}~\citep{tdmpc2}, a SOTA model-based RL algorithm that uses Model Predictive Control (MPC) to plan in the learned latent space of the model. 
Results are presented on three tasks -- PickCube, PushCube, and PullCube. 
For ALDA-DP, we compare against \textbf{Diffusion Policy}~\citep{diffusion_policy} and \textbf{Action Chunking with Transformers}~\citep{act} (ACT) on PickCube, PushCube, and the more difficult, long-horizon task PushT.
All methods are evaluated on the six visual variations shown in Figure~\ref{fig:mvgb} for their respective tasks. 
For each variation, we compute the average success rate over 1000 rollouts for the RL methods.
Due to computational constraints, we computed success rates over 500 rollouts for the BC methods. 
The results are presented in Figure~\ref{fig:main_plots}.

On PickCube, ALDA-SAC performs the best by significant margins on all visual variations, followed by ALDA-DP. 
On PushCube and PullCube, ALDA-SAC once again outperforms RL baselines by large margins. 
BC methods tend to outperform their RL counterparts on PushCube, where ALDA-DP and DP perform roughly the same, followed by ACT. 
We suspect that a combination of action-chunking, i.e., executing a sequence of actions before observing the next state, combined with the simplicity of the non-prehensile pushing task, makes unaltered Behavior Cloning baselines fairly robust to visual perturbations without requiring structured representations or associative memory. 
However, when fine-grained and precise manipulation is required, as with PickCube, visual randomizations present more of a challenge and are where ALDA-DP demonstrates superior performance.
The performance across ALDA- methods and baselines on PushT is low even without visual perturbations. 
Nonetheless, we find that ALDA-DP outperforms baselines on most visual variations, suggesting that a future, more powerful underlying BC algorithm that performs better on the base task will benefit from the structured representations that ALDA provides. 

\subsection{Equivariant Adaptation Results}
We finetune ALDA-DP and ALDA-SAC using the equivariant adaptation technique (Section~\ref{subsec:equi_adapt_method}) on three discrete cyclic groups, $C_8$, $C_{12}$, and $C_{24}$, i.e., three separate finetuned models.
These correspond to image rotations in 45, 30, and 15 degree increments, respectively. 
We train both models for 500 iterations, which takes at most 7 minutes for ALDA-DP on $C_{24}$, and 15 minutes for ALDA-SAC on $C_{24}$ on an RTX 3090. 
Evaluation on $C_n$ implies $n$ different image rotations, and for each rotation, we evaluate the finetuned model over 100 parallel environments and compute the average success rate. 
The average over all rotations' success rates on the PickCube task is in Table~\ref{tab:equiv_results}.

\begin{table}[h]
    \centering
    \resizebox{\textwidth}{!}{
    \begin{tabular}{l|c|c|c|c}
    \hline
        \textbf{Image Rotations} & \textbf{ALDA-SAC} & \textbf{ALDA-SAC (no finetune)} & \textbf{ALDA-DP} & \textbf{ALDA-DP (no finetune)}  \\
         \hline \textbf{}
         None &                           98.00 & 98.00 & 79.69  & 79.69 \\
         $C_8$ ($\Delta 45$ degrees) &    \textbf{97.44} &  1.78  & \textbf{76.86} & 3.79\\
         $C_{12}$ ($\Delta 30$ degrees) & \textbf{96.12} &  1.71  & \textbf{60.94} & 4.19 \\
         $C_{24}$ ($\Delta 15$ degrees) & \textbf{96.05} &  1.60  &\textbf{45.57} & 3.16 \\
    \hline
    \end{tabular}\textbf{}
    }
    \vspace{0.05in}
    \caption{Results of ALDA models finetuned using the equivariant adaptation technique on the PickCube task. For Group $n$, image observations are rotated every $\frac{360}{n}$ degrees, and success rates are calculated across 100 parallel environments. The average success rate over all rotations in the cyclic group is presented here.}
    \label{tab:equiv_results}
    \vspace{-0.2in}
\end{table}

ALDA-SAC performs the best and maintains high success rates, even with larger values of $N$. 
While ALDA-DP benefits from the finetuning procedure, we notice a drop-off in performance as the rotational discretization becomes more fine-grained. 
We hypothesize ALDA-DP is more challenging to finetune due to the larger model size, and will likely benefit from more iterations and/or deeper canonicalization networks. 
Nonetheless, our finetuning procedure results in strong robustness to image rotations, which will significantly aid in mitigating the effects of camera perturbations during real-world deployment. 

\subsection{Real-World Experiments}

\begin{table}[h]
    \centering
    \resizebox{\textwidth}{!}{
    \begin{tabular}{lccccc}
    \hline
    \textbf{Algorithm} & \textbf{Basic} & \textbf{Directed Light (Left/Middle/Right)} & \textbf{Ambient Light} & \textbf{Gray Cube} & \textbf{Distracting Objects} \\
     \hline
         ALDA Diffusion Policy & 80.0 & (\textbf{70.0} / 0.0 / 0.0) & \textbf{85.0} & \textbf{30.0} & \textbf{60.0} \\
         Diffusion Policy & 80.0 & (0.0 / 0.0 / 0.0) & 0.0 & 0.0 & 0.0 \\
         ACT & 80.0 & (55.0 / 0.0 / 0.0) & 15.0 & 0.0 & 0.0 \\
    \hline
    \end{tabular}
    }
    \vspace{0.05in}
    \caption{Average success rate of ALDA-DP, DP, and ACT on the PickCube task with the Franka Emika Panda arm under various visual perturbations. Results are averaged over 20 trials.}
    \label{tab:real_results}
    \vspace{-0.2in}
\end{table}

We collected 200 demonstrations via teleoperation of the Franka Emika Panda arm picking up a red cube, and used this dataset to train ALDA-DP, DP, and ACT. 
For evaluating the models, we designed 6 visual perturbation experiments and recorded the average success rate of each method over 20 trials. 
The "directed light" perturbations involve shining a flashlight on the workspace at three different angles -- from the left, middle, and right of the workspace. 
For the "ambient light" randomization, we turn on the overhead lights in the workspace, resulting in shadows of the arm and tabletop objects being cast onto the table. 
To test robustness to color randomizations, we tried picking up a gray cube instead of a red one. 
Finally, for the "distracting objects" randomization, we place various objects in the workspace and the background. 
Visualizations of these perturbations are presented in Figure~\ref{fig:main_fig}, and the results are presented in Table~\ref{tab:real_results}.
With the exception of ACT achieving decent performance on Directed Light (Left), ALDA-DP outperforms all baselines by significant margins. 
Directed Light Middle and Right proved to be too difficult for ALDA-DP, since the flashlight beam can cause visual occlusions of the cube and change its perceived color.


\section{Conclusion}
\label{sec:conclusion}
We present strong empirical evidence that disentangled representations, when paired with associative latent dynamics, enable robust zero-shot visual generalization for complex manipulation tasks in both simulation and real-world settings.
Our model achieves this without data augmentation, domain randomization, or camera calibration, and is robust to certain camera perturbations.
Outperforming SOTA RL and BC baselines, this approach marks meaningful progress toward generalist agents capable of solving tasks in the wild and provides an alternative to typical generalization paradigms to the robot learning community.
While the work here makes significant strides, much remains to be done in improving generalization performance.
For one, data diversity and data scaling performance are without question important pillars of developing generalist agents, and it remains to be seen if and how these techniques can be scaled to large models on massive datasets. 
Nonetheless, we believe the work here is a step in the right direction, and the remaining challenges present exciting opportunities for further research.

\section{Limitations}
We attempted several visual perturbations for which our approach was unsuccessful.
In simulation, changing the table color at evaluation time led to a near-total performance collapse of the model.
We suspect that without training data where the table and object colors are randomized independently, it is possible that the latent disentanglement procedure fails to separate object color and table color as independent sources of variation. 
Thus, changing one during evaluation can change the other and affect the downstream performance of the policy. 
Indeed, randomizing the table color during training mitigates the performance collapse, validating that data diversity is still an important component even when learning structured representations. 
During real-world evaluations, lifting the black curtains in the background (see Figure~\ref{fig:main_fig}) causes sunlight to bleed into the camera frame, resulting in an over-saturated white background that also causes performance collapse. 
While ALDA agents demonstrate robustness to distracting backgrounds, it seems that certain lighting conditions, as we also noticed with directed middle and right lighting (Table~\ref{tab:real_results}), remain challenging for our method. 
This could perhaps be mitigated with camera exposure balancing, and we remain optimistic that, with the right amount of exposure tuning, our method will zero-shot adapt to other scenes, such as other labs and rooms. 
Nonetheless, these negative results show that there is still much room for improvement.

\acknowledgments{We would like to thank Satyajeet Das and Karkala Hegde for assisting with the real world experiments, and Yigit Korkmaz for many insightful discussions.}

\clearpage


\bibliography{refs}  

\appendix

\newpage

\section{Latent Traversals}
\label{appendix:latent_traversals}

\begin{figure}[h]
    \centering
    \includegraphics[width=1.0\linewidth]{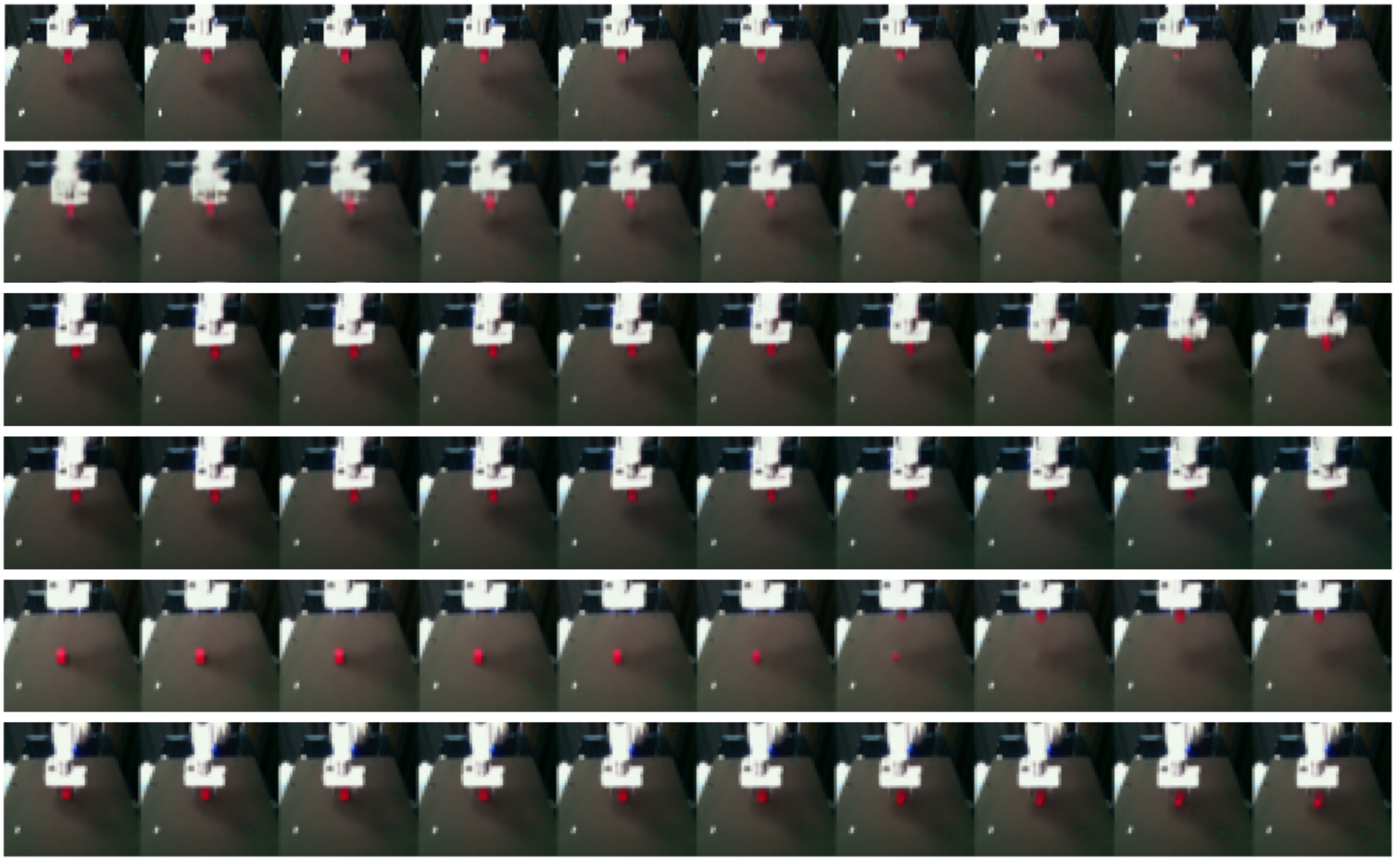}
    \caption{Traversing the disentangled latent space of ALDA-DP trained on real demonstrations and visualizing their corresponding reconstructions. Rows correspond to latent traversals of a single latent dimension given a reference image. By editing specific latent dimensions, we can visualize what factor of variation they correspond to. }
    \label{fig:latent_traversals}
\end{figure}

Since there are no quantitative metrics that tell us how well the representation disentangles without knowing the ground truth sources of variation a priori, following \cite{alda}, we present qualitative results instead. 
In this "latent traversal" experiment, we sample a batch of image observations from the expert-demonstrations dataset collected on the real Franka arm and encode them into disentangled representations using a trained ALDA-DP model. 
From here, we randomly sample an image from the batch and edit a randomly chosen dimension of the disentangled representation, interpolating it from [-1, 1]. 
The modified latent code is then passed to the decoder for visualizing the reconstructed image. 
Each row corresponds to the resulting reconstructed images from editing and interpolating a single latent dimension given a reference observation. 
The different rows correspond to latent traversals of different reference images. 
This experiment allows us to qualitatively see if editing a single latent dimension corresponds to a singular change in the resulting image, while also determining semantically what factor of variation it learns to represent. 

From Figure~\ref{fig:latent_traversals}, we find that indeed one latent dimension seems to correspond to a single factor of variation. 
For example, in the last row, the traversal of this latent dimension corresponds to different cube positions along the x-axis. 
Interestingly, several of the latent traversals show discontinuities, such as the second to last row where the cube is either on the table or picked up by the gripper. 
We hypothesize two reasons for this: (1) the discrete nature of the disentangled latent representation results in discontinuities, and (2) the task gradients from the BC objective encourage the agent to only pay attention to the cube either when it has yet to be picked up, or when it is already picked up and moved to the goal position, since these "critical" states potentially matter more than intermediate states. 
Future research into associative latent disentanglement models with continuous representations and/or representations that are more temporally consistent will greatly aid in certain tasks such as those with contact-rich dynamics or that require precise control. 

\section{Details on Associative Latent DisentAnglement}
\label{appendix:alda}

We first provide a formal definition of disentanglement. 
Since we wish to learn a representation such that each dimension of the latent embedding corresponds to a single factor of variation using a nonlinear model (e.g. neural network), the disentangled representation learning problem is often formulated as one of Nonlinear Independent Component Analysis, or \textbf{Nonlinear ICA}.
Suppose we are given a dataset of images $\mathcal{D} = \{\textbf{o}_1, \textbf{o}_2, ..., \textbf{o}_N \}$ with $n_s$ nonlinear independent source variables $s_1, ..., s_{n_s}$ that account for all variations within the data distribution. 
A hidden nonlinear generator function $g: \mathcal{S} \rightarrow \mathcal{O}$ maps the sources to the observations:

\begin{equation*}
    p(\textbf{s}) = \prod_{i=1}^{n_s} p(s_i), \textbf{o} = g(\textbf{s}).
\end{equation*}

The goal of Nonlinear ICA and thus disentangled representation learning is to uncover the hidden sources $s_1, ..., s_{n_s}$ factorized from each other. 

Under the hood, ALDA employs an adaptation of QLAE~\citep{qlae} to disentangle the latent representation. 
An image observation $\textbf{o}_t$ is first encoded into a continuous representation $\textbf{z}_{cont.} \in \mathbb{R}^{n_z}$, where $n_z$ is the number of independent sources of variation that form the basis of the observation distribution. 
Each dimension $z_j, j = 1,..., n_z$ of $\textbf{z}_{cont.}$ is mapped to a discrete value by a collection of scalar codebooks $Z = V_1 \times V_2 \times ... \times V_{n_z}$, one per latent dimension via attention-based association: 

\begin{equation*}
    z_{d_j} = \text{Softmax} \left( \beta \text{Sim}(z_j, V_j) \right) \odot V_j
\end{equation*}

where $\text{Sim}(\cdot, \cdot)$ is any similarity function and $\beta$ is a hyperparameter that controls the degree of separation between latent values. 
We use the negative $L_1$ distance as our similarity function:
The disentangled latent representation is used to reconstruct the original observation as an auxiliary objective using a reconstruction loss $\mathcal{L}_{recon.}$
Although gradients can flow through the Softmax operator, in practice and with large values of $\beta$, the gradients can become extremely large and destabilize training, so we instead use a StopGrad operator and optimize the encoder to be close to the discretized latent distribution using a commitment loss term: 

\begin{equation*}
    \mathcal{L}_{commit.} = || \text{StopGrad}(\textbf{z}_{cont.}) - \textbf{z}_d||_2^2 \text{.}
\end{equation*}

Large activation penalties $\lambda_{\theta}||\theta||_2^2, \lambda_{\phi}||\phi||_2^2$ are applied to the encoder and decoder weights $\theta \text{ and } \phi$ serving as an information bottleneck which, inline with prior disentanglement methods~\citep{fsq, bVAE, bio_ae}, facilitates the disentanglement process.
Put together, the ALDA objective is 

\begin{equation*}
    J(ALDA) = \mathbb{E}_{\textbf{o}_t \sim \mathcal{D}} \big[ 
        w_1\mathcal{L}_{recon.} + w_2\mathcal{L}_{commit} + \lambda_{\theta}||\theta||_2^2 + \lambda_{\phi}||\phi||_2^2
    \big].
\end{equation*}
For all experiments, we set $\lambda_{\phi} \text{ and } \lambda_{\theta} \text{ to } 0.1$ and $w_1, w_2$ to 1.0 and 0.1, respectively. 

\newpage

\section{ALDA-SAC}

\begin{figure}[h]
    \centering
    \includegraphics[width=1.0\linewidth]{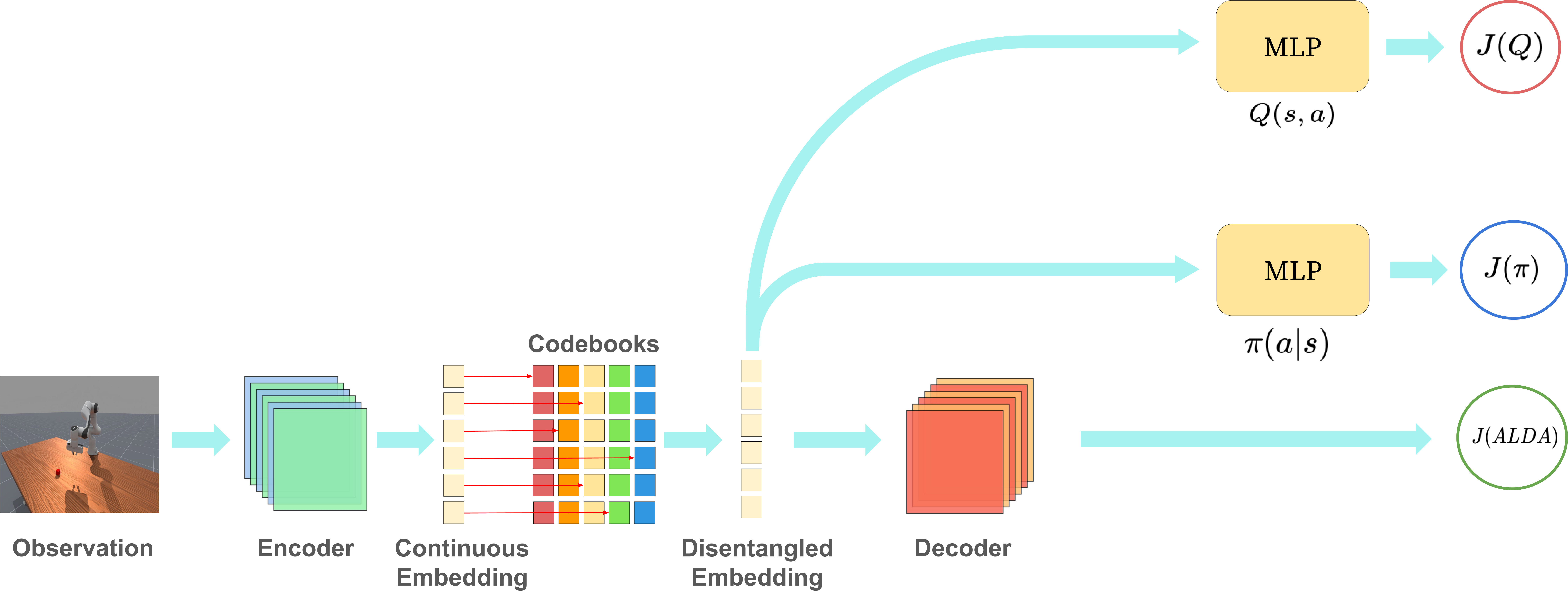}
    \caption{ALDA-SAC method diagram.}
    \label{fig:alda_sac}
\end{figure}

The original ALDA-SAC implementation in \cite{alda} assumes as input a stack of observations $\in \mathbb{R}^{k \times C \times H \times W}$.
The ALDA model is trained to independently encode and reconstruct each individual frame and disentangle the latent representation according to the objective $J(ALDA)$ (Section \ref{appendix:alda}).
The stack of latent states $\in \mathbb{R}^{k \times |z_d|}$ is fed into a 1D-CNN to extract temporal information before being input into a linear projection layer and then finally the actor/critic networks.
However, we did not find frame stacking or projection to be beneficial on the ManiSkill3 manipulation tasks, and instead opt for a simplified implementation, illustrated in Figure \ref{fig:alda_sac}.
Here, a single observation is disentangled into a latent representation and is input directly into the actor/critic networks, which are trained according to the policy and critic objectives as in standard Soft Actor-Critic~\citep{sac}. 

\newpage 

\section{Equivariant Adaptation of Policies -- Algorithm Pseudocode}
\label{appendix:equi_adapt}

\begin{algorithm}
    \caption{Equivariant Adaptation}
    \label{alg:equi_adapt}
    \begin{algorithmic}
        \STATE {\bfseries Input:} Pretrained agent $\mathcal{A}_\theta$ (encoder $f$, latent model $l$, policy $\pi$), replay buffer $\mathcal{D}$, "oracle" clone of the pretrained agent with frozen weights $\mathcal{A}^*$, canonicalizer $C_{\phi}$ i.e. an equivariant CNN, $\mathcal{L}_{prior}$ hyperparameter $\beta$, learning rate $\alpha$, training steps $N$.
        \STATE
        \FOR{$i \leftarrow 1 \; \textbf{to} \; N$}
            \STATE $\textbf{o} \in \mathbb{R}^{B \times C \times H \times W} \sim \mathcal{D}$
            \STATE $\textbf{o}^{canon} \leftarrow C_{\phi}(\textbf{o})$
            \STATE $a \leftarrow \mathcal{A}_{\theta}(\textbf{o}^{canon})$
            \STATE $a^* \leftarrow \mathcal{A}^*(\textbf{o}) $
            \STATE $z_{cont.}^{canon} \leftarrow f(\textbf{o}^{canon})$
            \STATE $z_{d}^{canon} \leftarrow l(z_{cont.}^{canon})$
            \STATE $\mathcal{L}_{act} \leftarrow ||a - a^*||_2^2$
            \STATE $\mathcal{L}_{prior} \leftarrow - D_{KL} (\mathbb{P}_D || P_{C(\textbf{o})})$
            \STATE $\mathcal{L}_{commit} = ||\text{StopGrad}(z_{d}^{canon}) - z_{cont.}^{canon}||_2^2$
            \STATE $\mathcal{L}_{total} = \mathcal{L}_{act} + \beta \mathcal{L}_{prior} + \mathcal{L}_{commit}$
            \STATE $\nabla \theta \leftarrow \frac{\partial \mathcal{L}_{total}}{\partial \theta}$
            \STATE $\theta \leftarrow \theta + \alpha \nabla \theta$
            \STATE $\nabla \phi \leftarrow \frac{\partial \mathcal{L}_{total}}{\partial \phi}$
            \STATE $\phi \leftarrow \phi + \alpha \nabla \phi$
        \ENDFOR
    \end{algorithmic}
\end{algorithm}

Algorithm \ref{alg:equi_adapt} contains pseudocode for the equivariant adaptation of a pretrained robot policy. 
A lightweight ENN $C_{\phi}$ for the discrete SO(2) symmetry group $C_n$ is initialized and trained to canonicalize the input observation image \textbf{o}, which may be rotated by an arbitrary angle $ \frac{360}{n} \cdot i, i\in[0,n]$. 
The pretrained policy $\mathcal{A}_{\theta}$ is jointly finetuned to produce optimal actions by minimizing the error between its outputs given the canonicalized observation and the outputs of a cloned version of itself with frozen weights given the original observation. 

\newpage

\section{Implementation Details}

\begin{table}[h]
    \centering
    \begin{tabular}{|c|c|}
    \hline
        \textbf{Parameter} & \textbf{Value}  \\
        \hline
            Learning rate & 1e-4 \\
            Obs horizon & 2 \\ 
            Action horizon & 8 \\
            Prediction horizon & 16 \\
            Diffusion Embedding Dim & 64 \\
            Training steps & 3e5 \\
            Image resolution & 64 \\
            Number of latents $|z_d|$ & 20 \\
            Values per latent $|V|$ & 20 \\
            \hline
    \end{tabular}
    \caption{Task agnostic hyperparameters for ALDA-DP.}
    \label{tab:hyperparams}
\end{table}

\begin{table}[h]
    \centering
    \begin{tabular}{|l|c|c|c|}
        \hline
         \textbf{Task} & \textbf{\# Demonstrations} & \textbf{Episode Length} & \textbf{Camera View} \\
         \hline
         PickCube & 1000 & 100 & Angled \\
         PushCube & 997 & 100 & Front \\
         PushT & 800 & 200 & Front \\
         \hline
    \end{tabular}
    \caption{Task specific parameters for ALDA-DP's simulation results.}
    \label{tab:task_params}
\end{table}

Task-agnostic and task-specific hyperparameters for ALDA-DP are given in tables \ref{tab:hyperparams} and \ref{tab:task_params}, respectively. 
For a list of ALDA-SAC hyperparameters, we refer the readers to \cite{alda}. 
We use largely the same hyperparameters, with the following exceptions: we do not incorporate framestacking, and the "number of latents" and "values per latent" parameters are set to 10 and 12, respectively. 

\subsection{Camera View}
ManiSkill3 defines the camera location and orientation using the "look at" convention, which accepts as arguments the 3D position of the camera, and the 3D "target" position i.e. where the camera is "looking" with respect to the world frame.
For tabletop manipulation tasks, the world frame is roughly the center of the workspace and axis-aligned with the table. 
By default, these parameters are set to $(0.3, 0.0, 0.6)$ and $(-0.1, 0.0, 0.1)$, i.e. the camera is elevated, front-facing, and pointed downwards roughly towards the center of the workspace. 
However, this camera view mostly captures the table and very little of the surrounding background, making the "DistractingBackground" visual perturbation less challenging, especially on the PickCube task where we noticed background randomizations having the largest impact on performance. 
Thus, we define a new "angled" view with the camera and target positions set to $(0.4, 0.5, 0.6)$ and $(0.0, 0.0, 0.35)$, which we use for the PickCube task. 
This makes the DistractingBackground perturbation more challenging for all methods while maintaining full view of the workspace. 

\subsection{Real-World Setup}
We use a Franka Emika Panda arm, a RealSense D515 camera, and a 3D printed red cube for our main experiments. 
200 real demonstrations were collected via teleoperation under visually consistent conditions to train the ALDA-DP model. 
Proprioceptive state is a 20-dim vector consisting of the arm's joint angles $\in \mathbb{R}^7$, gripper width$\times2$, an "is grasped" boolean $\in \{0, 1\}$, the end-effector pose $\in \mathbb{R}^3$ and quaternion rotation $\in \mathbb{R}^4$, and the goal position $\in \mathbb{R}^3$. 
Actions are a 4-dimensional vector consisting of the target end effector position $\in \mathbb{R}^3$ and a binary gripper open/close command $\in \{0, 1\}$.
This largely mimics the state and action information given by ManiSkill3, except that we do not include joint velocities when training for real-world deployment. 
All other hyperparameters are the same as the ALDA-DP model trained in simulation.

\end{document}